\documentclass{article}

\usepackage{macros}
\usepackage{natbib}
\usepackage{wrapfig}

\title{Mollifying Networks}

\author{
    Caglar Gulcehre$^1$, \  Marcin Moczulski$^{2, \ast}$, \  Francesco Visin$^{3, \ast}$ \\
  {\bf Yoshua Bengio}$^1$ \\
  $^1$ University of Montreal, $^2$ University of Oxford,  $^3$ Politecnico di Milano \\
}

\begin{document}

\maketitle
\begin{abstract}
    The optimization of deep neural networks can be more challenging than
    traditional convex optimization problems due to the highly non-convex
    nature of the loss function, e.g. it can involve pathological landscapes
    such as saddle-surfaces that can be difficult to escape for algorithms
    based on simple gradient descent. In this paper, we attack the problem of
    optimization of highly non-convex neural networks by starting with a
    smoothed -- or \textit{mollified} -- objective function which becomes more
    complex as the training proceeds.  Our proposition is inspired by the recent
    studies in continuation methods: similar to curriculum methods, we begin
    learning an easier (possibly convex) objective function and let it evolve
    during the training, until it eventually goes back to being the original,
    difficult to optimize, objective function. The complexity of the mollified
    networks is controlled by a single hyperparameter which is annealed during 
    the training. We show improvements on various difficult optimization tasks and establish a
    relationship between recent works on continuation methods for neural networks
    and mollifiers.
\end{abstract}
\blfootnote{$^\ast$ This work was done while these students were interning at
    the MILA lab. at the University of Montreal}
\vspace*{-1mm}
\section{Introduction}
\vspace*{-1mm}

Deep neural networks -- i.e. convolutional networks \citep{LeCun:1989:BAH:1351079.1351090},
LSTMs \citep{Hochreiter+Schmidhuber-1997} or GRUs \citep{cho2014learning} --
achieve state of the art results on a range of challenging tasks like
object classification and detection~\citep{Szegedy2014}, semantic segmentation~\citep{visin2015reseg},
speech recognition \citep{38131}, statistical machine translation
\citep{sutskever2014sequence,bahdanau2014neural}, playing Atari
\citep{Deepmind-atari-arxiv2013} and Go \citep{silver2016mastering}.
When trained with variants of SGD \citep{Bottou98} deep models can be hard to
optimize due to their highly non-linear and nonconvex nature \citep{Choromanska-et-al-AISTATS2015,Dauphin-et-al-NIPS2014-small}.

A number of approaches were proposed to alleviate the difficulty of optimization:
addressing the problem of the internal covariate shift with Batch Normalization \citep{DBLP:journals/corr/IoffeS15},
learning with a curriculum \citep{bengio2009curriculum} and recently training with
diffusion \citep{mobahi2016training} - a form of continuation method.
At the same time, the impact of noise injection on the behavior of modern deep models has been
explored in \citep{DBLP:journals/corr/NeelakantanVLSK15} and it has been recently shown
that noisy activation functions improve performance on a wide variety of tasks \citep{gulcehre2016noisy}.

In this paper, we connect the ideas of curriculum learning and
continuation methods with those arising from models with skip connections and
with layers that compute near-identity transformations.  Skip connections allow
to train very deep residual and highway architectures
\citep{he2015deep,srivastava2015training} by skipping layers or block of
layers. Similarly, it is now well known that it is possible to stochastically
change the depth of a network during training
\citep{DBLP:journals/corr/HuangSLSW16} and still converge.

In this work, we introduce the idea of mollification -- a form of
differentiable smoothing of the loss function connected to noisy activations --
which can be interpreted as a form 
adaptive noise injection that only depends on a single hyperparameter. Inspired
by \citet{DBLP:journals/corr/HuangSLSW16}, we exploit the same hyperparameter
to stochastically control the depth of our network. This allows us to start the
optimization in the easier setting of a \textbf{convex} objective function (as long as
the optimized criterion is convex, e.g. linear regression, logistic regression)
and to slowly introduce more complexity into the model by annealing the
hyperparameter making the network deeper and increasingly non-linear.

\section{Mollifying Objective Functions}
\label{sec:moll_obj_func}
In this section we first describe continuation and annealing methods, we then
introduce mollifiers and show how they can be used to ease the optimization as
a continuation method that gradually reduces the amount of smoothing applied to
the training objective of a neural network.

\vspace*{-1mm}
\subsection{Continuation and Annealing Methods}
\vspace*{-1mm}

\begin{figure}
    \vspace*{-2mm}
    \centerline{\resizebox{0.4\textwidth}{!}{\includegraphics{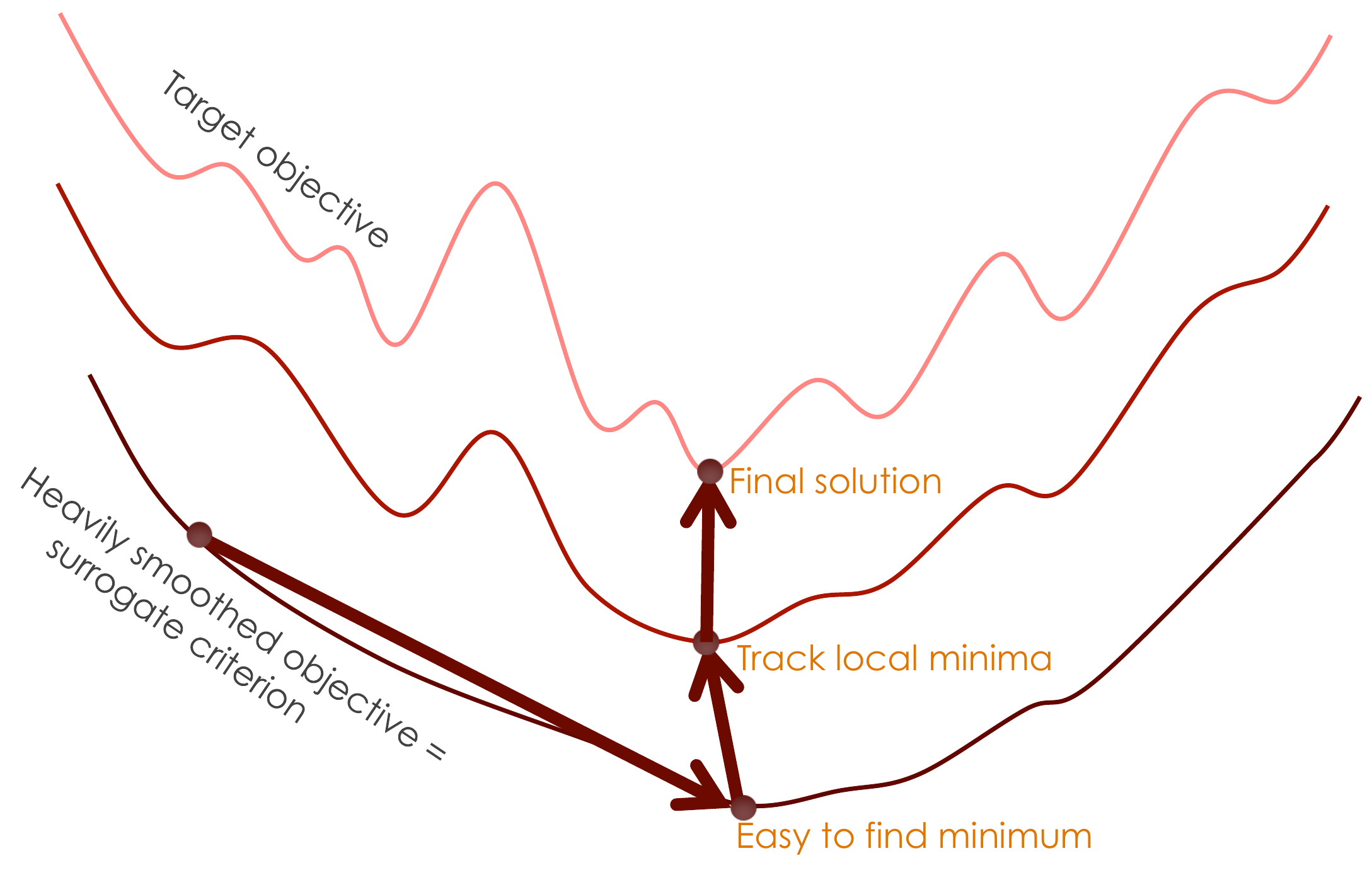}}}
      \vspace*{-1mm}
    \caption{A sequence of optimization problems of increasing complexity, where
        the first ones are easy to solve but only the last one corresponds to the
        actual problem of interest. It is possible to tackle the problems in order,
        starting each time at the solution of the previous one and tracking the
        local minima along the way.
    }
    \label{fig:continuation}
\end{figure}
\vspace*{-2mm}

{\em Continuation methods} and {\em simulated annealing} provide a general
strategy to reduce the impact of local minima and deal with non-convex,
continuous, but not necessarily everywhere differentiable objective functions.

Continuation methods~\citep{Allgower80-short}, address a complex optimization
problem by smoothing the original function, turning it into a different problem
that is easier to optimize. By gradually reducing the amount of smoothing, it
is possible to consider a sequence of optimization problems that converge to
the optimization problem of interest (see Fig.~\ref{fig:continuation}).


These methods have been very successful for tackling difficult optimization
problems involving non-convex objective functions with multiple local minima
and possibly points of non-differentiability. In machine learning, approaches
based on curriculum learning~\citep{bengio2009curriculum} are inspired by this
principle to define a sequence of gradually more difficult training tasks (or
training distributions) that converge to the task of interest. Gradient-based
optimization over a sequence of mollified objective functions has been shown to
converge~\citep{Chen-2012}.

In the context of stochastic gradient descent, we can use an estimator of the
gradient of the smoothed objective function. This is convenient because
actually computing the smoothed objective function may not be analytically
feasible, but a Monte-Carlo estimate can often be obtained easily.

\begin{definition}{\textbf{Weak gradients (Distributional Gradients)}}
     We generalize the definition of weak/distributional derivatives to
     gradients in order to show the relationship with training neural networks.
     For an integrable function $\LL$ in space $\LL \in L([a, b])$, $\g \in L([a, b]^n)$ is
     a $n\emph{-}dimensional$ weak gradient of $\LL$ if it satisfies the Eq. \ref{eqn:weak_derivative}:
 \begin{equation}
     \label{eqn:weak_derivative}
     \int_C \g(\TT) K(\TT) d\TT = -\int_C \LL(\TT)\nabla K(\TT) d\TT,
 \end{equation}
 where $K(\TT)$ is an infinitely differentiable function vanishing at infinity and $C
 \in [a, b]^n$, $\TT \in \R^{n}$ is a parameter vector and
 $\LL(\TT)$ is the cost function that we would like to minimize.
 \end{definition}

 \begin{definition}{\textbf{Mollifiers}}
A mollifier is an infinitely differentiable function that behaves as an approximate identity in
the group of convolutions of integrable functions. $K$ is a mollifier if it is
infinitely differentiable and for any integrable function $\LL$ we have:
\begin{align*}
    \LL(\TT) = \lim_{\epsilon\to0} \int \epsilon^{-1}K(\vx/\epsilon) \LL(\TT-\vx) d\vx .
\end{align*}
\end{definition}
A mollifier converges to the Dirac function if we rescale it appropriately. In
general, we are interested in constructing a {\em sequence of mollifiers}
indexed by $\epsilon$, which in the above integral corresponds to
$\epsilon^{-1}K(\vx/\epsilon)$.  This allows one to construct a sequence of
gradually less blurred and closer approximations to $\LL$.

We can define a weak gradient of a non-differentiable function by convolving it with a
mollifier \citep{evans1998partial}:
\begin{align*}
    \nabla (\LL \ast K)(\TT) = (\LL \ast \nabla K)(\TT) .
\end{align*}

We can choose $K(\cdot)$ to be the density function of an accessible distribution such as a Gaussian distribution.
In the limit, the gradient of a mollified function is equivalent to its mollified weak-gradients:
\begin{align*}
    \g(\TT) &= -\lim_{\epsilon\to0} \int \epsilon^{-1}K(\vx/\epsilon) \g(\TT-\vx) d\vx .\\
\end{align*}
From the properties of weak-gradients presented in Eq. \ref{eqn:weak_derivative}, we know that:
\begin{align*}
    \g(\TT) &= \lim_{\epsilon\to0} \int \epsilon^{-1}\nabla K(\vx/\epsilon) \LL(\TT-\vx) d\vx .
\end{align*}

An important property of mollifiers and weak-gradients is that they allow to
backpropagate through functions that don't have strong derivatives defined
everywhere, such as e.g. a step function.

We can obtain the mollified version $\LL_K(\TT)$ of the cost function $\LL(\TT)$ by convolving it
with a mollifier $K(\TT)$.  Similarly to the analysis in \citet{mobahi2016training}, we
can write a Monte-Carlo estimate of $\LL_K(\TT) = (\LL \ast K)(\TT) \approx \frac{1}{N}
\sum_{i=1}^{N} \LL(\TT-\xi^{(i)})$. We provide the derivation and the gradient of this equation in
Appendix \ref{sec:MonteCarloMolly}.  $K(\cdot)$ is the kernel which we mollify with and corresponds
to the average effect of injecting noise $\xi$ sampled from standard Normal distribution. The amount of noise
controls the amount of smoothing. Gradually reducing the noise during training
is related to a form of {\em simulated annealing}~\citep{Kirkpatrick83}.

This result can be easily extended to neural networks, where the layers
typically take the form:
\begin{equation}
	\vh^{l} = \f(\mW^{l}\vh^{l-1})
\end{equation}
with $\vh^{l-1}$ a vector of activations from the layer below, $\mW^{l}$ a
matrix representing a linear transformation and $\f$ an element-wise
non-linearity of choice.

A mollification of such a layer can be formulated as:
 \begin{equation}
         \vh^{l} = \f((\mW^{l} - \xi^{l})\vh^{l-1}) \text{, where } \xi^{l} \sim \mathcal{N}(\mu, \sigma^2)
   \label{eqn:noisy_layer}
 \end{equation}

%
%
\vspace*{-1mm}
\subsection{Generalized and Noisy Mollifiers}
\vspace*{-1mm}

We introduce a slight generalization of the concept of mollifiers that encompasses the approach
we explored here and that is targeted at optimization via a continuation method using stochastic
gradient descent.
%
%
\begin{definition}{\textbf{(Generalized Mollifier)}}.
A {\em generalized mollifier} is a transformation $T_\sigma(f)$ in $\R^{k}$ to $\R^{k}$ of a function $f$ such that:
\begin{equation}
   \lim_{\sigma\rightarrow 0} T_\sigma f = f,
  \label{eq:converges}
\end{equation}
\begin{equation}
   f^0 = \lim_{\sigma\rightarrow \infty} T_\sigma f \;\;\;{\rm is}\;{\rm convex}
  \label{eq:convex}
\end{equation}
\begin{equation}
 \frac{\partial (T_\sigma f)(x)}{\partial x}\;\;\; {\rm exists} \; \forall x, \sigma>0
  \label{eq:differentiable}
\end{equation}
\end{definition}

In addition we consider noisy mollifiers which can be defined as the
expected value of a stochastic function $\phi(x,\xi)$ under some noise source $\xi$ with variance $\sigma$:
\begin{equation}
  (T_\sigma f)(x) = E_\xi[\phi(x,\xi_{\sigma})]
  \label{eq:corruption}
  \end{equation}
\begin{definition}{\textbf{(Noisy Mollifier)}}.
  We call a stochastic function $\phi(x,\xi_{\sigma})$ with input $x$ and noise $\xi$ a {\em noisy mollifier} if its
  expected value corresponds to the application of a generalized mollifier $T_\sigma$,   as per Eq.~\ref{eq:corruption}.
\end{definition}

The composition of two noisy mollifiers sharing the same $\sigma$ is a noisy mollifier, since
the three properties in the definition (Eqs.~\ref{eq:converges},\ref{eq:convex},\ref{eq:differentiable})
are satisfied. When $\sigma=0$ no noise is injected and therefore the original
function is being optimized. If $\sigma\rightarrow\infty$ instead, the
function will become convex.

Consequently, corrupting separately the activation function of each level of a
deep neural network (but with a shared noise level $\sigma$) and annealing
$\sigma$ yields a noisy mollifier for the objective function. This is related
to the work of \citet{mobahi2016training}, who recently introduced analytic
smooths of neural network non-linearities in order to help training recurrent
networks. The differences with the work presented here are twofold:
we use a noisy mollifier (rather than an analytic smooth of the network
non-linearities) and we introduce (in the next section) a particular form of
the noisy mollifier that empirically proved to work very well.

\citet{mobahi2016training} also makes a link between continuation or annealing methods and noise
injection, although an earlier form of that observation was already made by~\citet{Bottou_these91}
in the context of gradually decreasing the learning rate when doing stochastic gradient descent.
The idea of injecting noise into a hard-saturating non-linearity was previously used
in~\citet{Bengio-arxiv2013} to help backpropagate signals through semi-hard decisions (with the
``noisy rectifier'' stochastic non-linearity).

\section{Method}

We focus on improving the optimization of neural networks with $\tanh(\cdot)$ and
$\text{sigmoid}(\cdot)$ activation functions, since those have been known to be
particularly challenging to optimize and play a crucial role in models that
involve gating (e.g.  LSTM, GRU) as well as piecewise linear activations such as ReLUs.
However the general principles we present in this paper can be easily
adapted to the other activation functions as well.

We propose a novel learning algorithm to mollify the cost of a neural network that addresses an
important drawback of previously proposed noisy training procedures: as the noise gets larger it
can dominate the learning process and lead the algorithm to perform a random walk on the energy
landscape of the objective function. Conversely in the algorithm we propose in this paper, as the
noise gets larger the SGD minimizes a simpler e.g. convex, but still meaningful objective
function. To this end we define the desired behavior of the network in the
limit cases where the noise is very large or very small, and modify the model
architecture accordingly.

In other words during training we minimize a sequence of noisy objectives
$\mathrm{L} = (\LL^{1}(\TT;\xi_{\sigma_1}), \LL^{2}(\TT;\xi_{\sigma_2}),
\cdots, \LL^{k}(\TT;\xi_{\sigma_k}))$ where the scale (variance) of the noise
$\sigma_i$ and the simplicity of the objective function will be reduced during
the training. Our model still satisfies the basic properties of the
\textit{generalized and noisy mollifiers}.

Instead of mollifying our objective function with a kernel, we propose to mimic
the properties of the mollification that are important for the continuation by
reformulating our objective function such that:

\begin{enumerate}
    \item We start the training by optimizing a convex objective function which is obtained by
        configuring all layers between the input and the last cost layer to compute an identity function.
        On the other hand, a high level of noise controlled with a single scalar $p \in [0, 1]$
        per layer assures that element-wise activation functions compute a linear function.
    \item As the noise level $p$ is annealed we move from identity transformations to
         arbitrary linear transformations between layers.
    \item Simultaneously the decreasing level of noise $p$ allows element-wise activation functions
    to become non-linear.
\end{enumerate}

On the other hand, this kind of noisy training is potentially helpful for the
generalization as well, since the noise in the noisy mollified model will make
the backpropagation to be noisy as well. Due to the noise induced by the
backpropagation through the noisy units \cite{hochreiter1997flat}, SGD is more
likely to converge to a flatter-minima because the noise will
help SGD escape from sharper local minima.

\subsection{Simplifying the Objective Function for Feedforward Networks}

\begin{figure}
\centerline{\resizebox{0.9\textwidth}{!}{\includegraphics{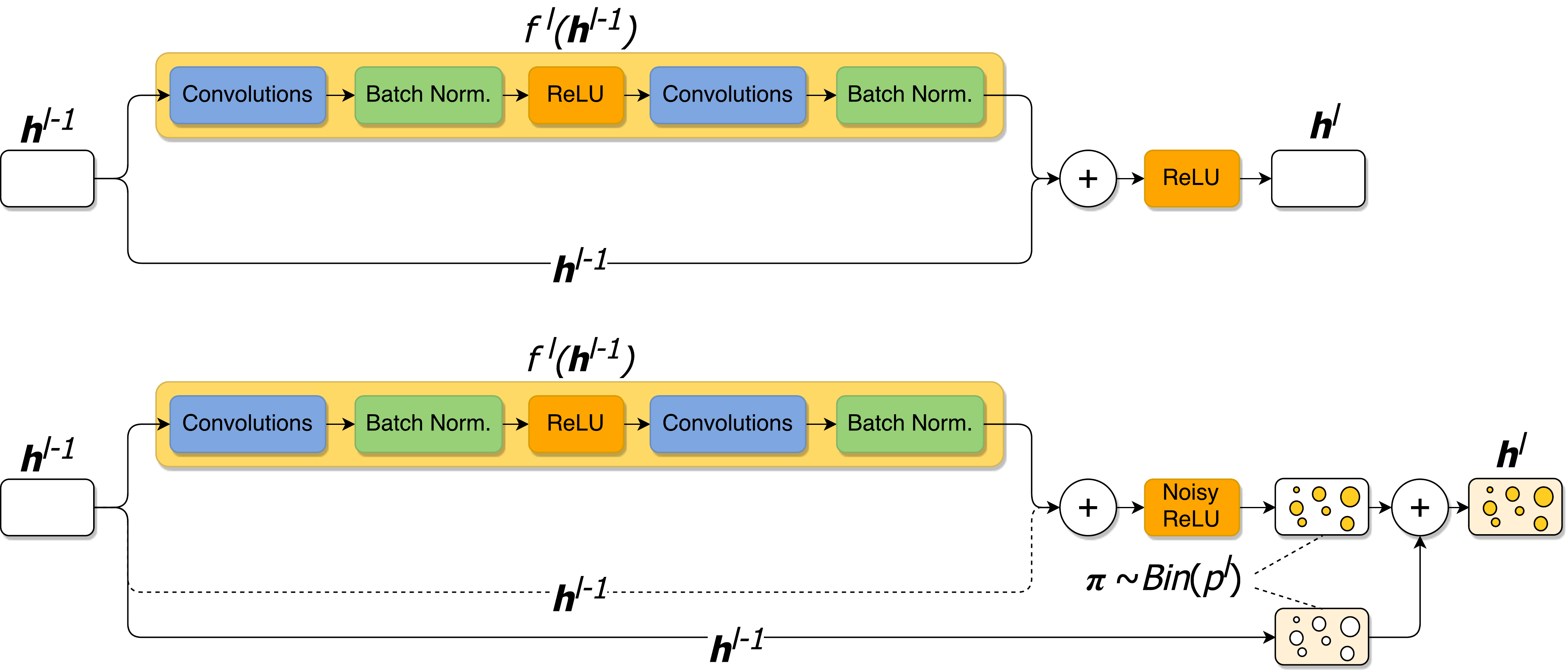}}}
\caption{\textbf{Top:} Stochastic depth. \textbf{Bottom:} mollifying network.
    The dashed line represents the optional residual connection. In the top
    path, the input is processed with a convolutional block followed by a noisy
    activation function, while in the bottom path the original activation of
    the layer $l-1$ is propagated untouched. For each unit, one of the two
    paths in picked according to a binary stochastic decision $\boldsymbol\pi$.
}
\label{fig:molly_layer}
\end{figure}

For \emph{every unit} of each layer, we either copy the activation (output) of
the corresponding unit of the previous layer (the identity path
in Figure~\ref{fig:molly_layer}) or output a noisy activation $\tilde{\vh}^{l}$
of a non-linear transformation of it $\psi(\vh^{l-1}, \boldsymbol\xi; \mW^l)$,
where $\boldsymbol\xi$ is noise, $\mW^l$ is a weight matrix applied on
$\vh^{l-1}$ and $\boldsymbol\pi$ is a vector of binary decisions for each unit
(the convolutional path in Figure~\ref{fig:molly_layer}):
\begin{align}
    \tilde{\vh}^{l} &= \psi(\vh^{l-1}, \boldsymbol\xi; \mW^l) \\
    \phi(\vh^{l-1}, \boldsymbol\xi, \boldsymbol\pi^l; \mW^l) &= \boldsymbol\pi^l \odot \vh^{l-1} + (1 - \boldsymbol\pi^l) \odot \tilde{\vh^l} \\
    \vh^{l} &= \phi(\vh^{l-1}, \boldsymbol\xi, \boldsymbol\pi^l; \mW^l).
    \label{eqn:psi_linear_conn_below}
\end{align}

To decide which path to take, for each unit in the network, a binary stochastic
decision is taken by drawing from a Binomial random variable with probability
dependent on the decaying value of $p^l$:
\begin{equation}
    \boldsymbol\pi^l \sim \operatorname{Bin}(p^l)
\label{eqn:bin_sampling}
\end{equation}

If the number of hidden units of layer $l-1$ and layer $l+1$ is not the same,
we can either zero-pad layer $l-1$ before feeding it into the next layer or
apply a linear projection to obtain the right dimensionality.

For $p^l = 1$, the layer computes the identity function leading to a \textbf{convex}
objective. If $p^l = 0$ the layer computes the original non-linear
transformation unfolding the full capacity of the model.


%
%
\paragraph{Linearizing the network}
In section \ref{sec:moll_obj_func}, we show that convolving the objective
function with a particular kernel can be approximated by adding noise to the
activation function. This method may suffer from excessive random exploration
when the noise is very large.

We address this issue by bounding the element-wise activation function
$\f(\cdot)$ with its linear approximation when the variance of the noise is
very large, after centering it at the origin. The resulting function
$\f^{\ast}(\cdot)$ is bounded and centered around the origin.
%
%
%
%
Note that centering the $\text{sigmoid}$ or $\text{hard-sigmoid}$ will make
them symmetric with respect to the origin.
%
With a proper choice of the standard deviation $\sigma(\vh)$, the noisy
activation function becomes a linear function  of the input when p is large, as
illustrated by Figure~\ref{fig:linearizing_with_noise}.

Let $\fu^{\ast}(x) = \fu(x) - \fu(0)$, where $\fu(0)$ is the offset of the function
from the origin, and $x_i$ the $i$-th dimension of an affine transformation of the
output of the previous layer $\vh^{l-1}$: $x_i = \vw_i^{\top} \vh^{l-1} + b_i$.
Then:
%
%
\begin{equation}
    \psi(x_i, \xi_i; \vw_i) = \sgn(\fu^{\ast}(x_i))
    \text{min}(|\fu^{\ast}(x_i)|, |\f^{\ast}(x_i) +
    \sgn(\fu^{\ast}(x_i)) |s_i||) + \fu(0)
\end{equation}

\begin{figure}[htp]
    \centering
    \begin{minipage}{.4\textwidth}
       \centering
       \includegraphics[width=\textwidth]{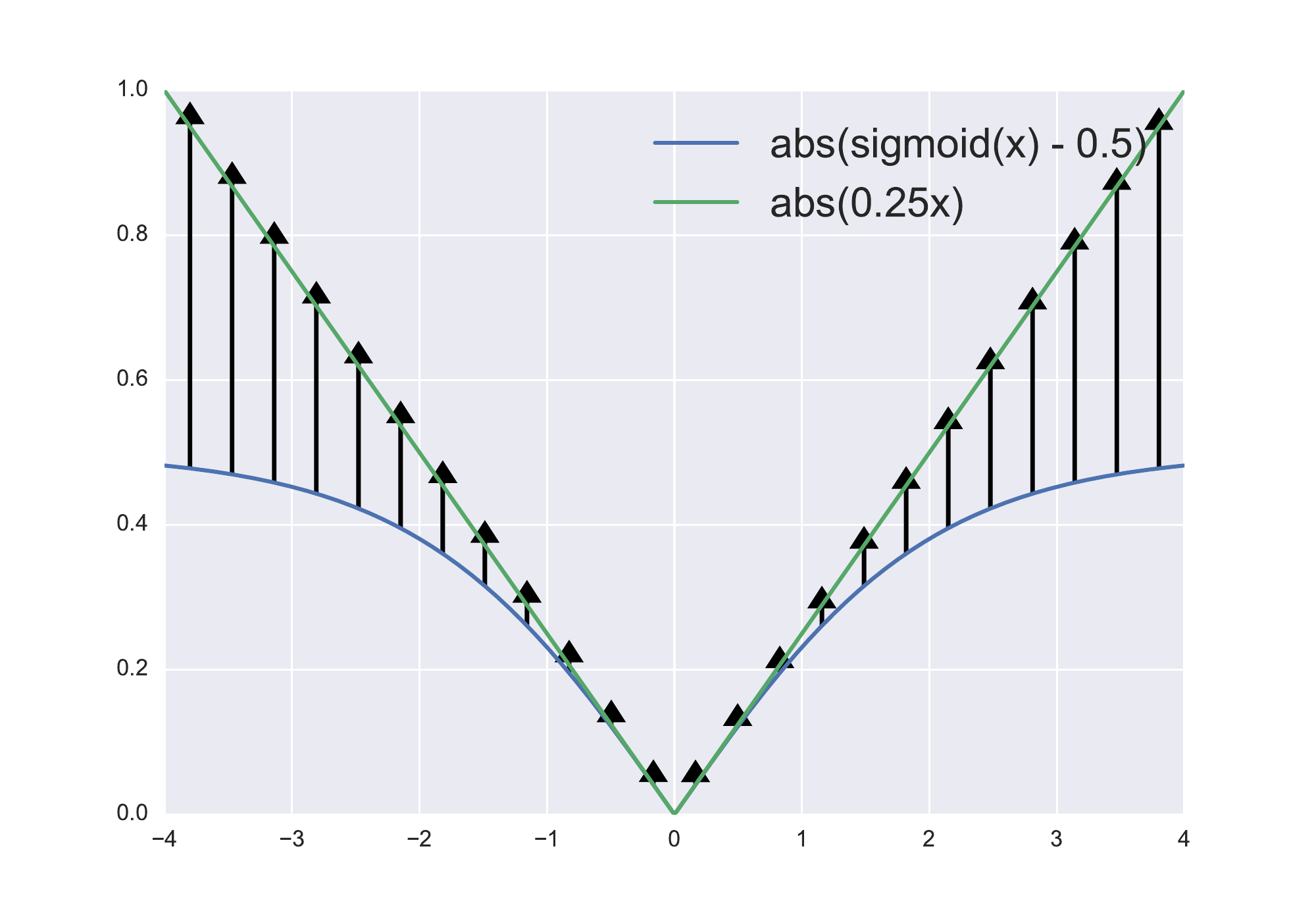}
       \caption*{a)}
       \label{fig:test1}
     \end{minipage}
     \hfil
     \begin{minipage}{.4\textwidth}
        \centering
        \includegraphics[width=\textwidth]{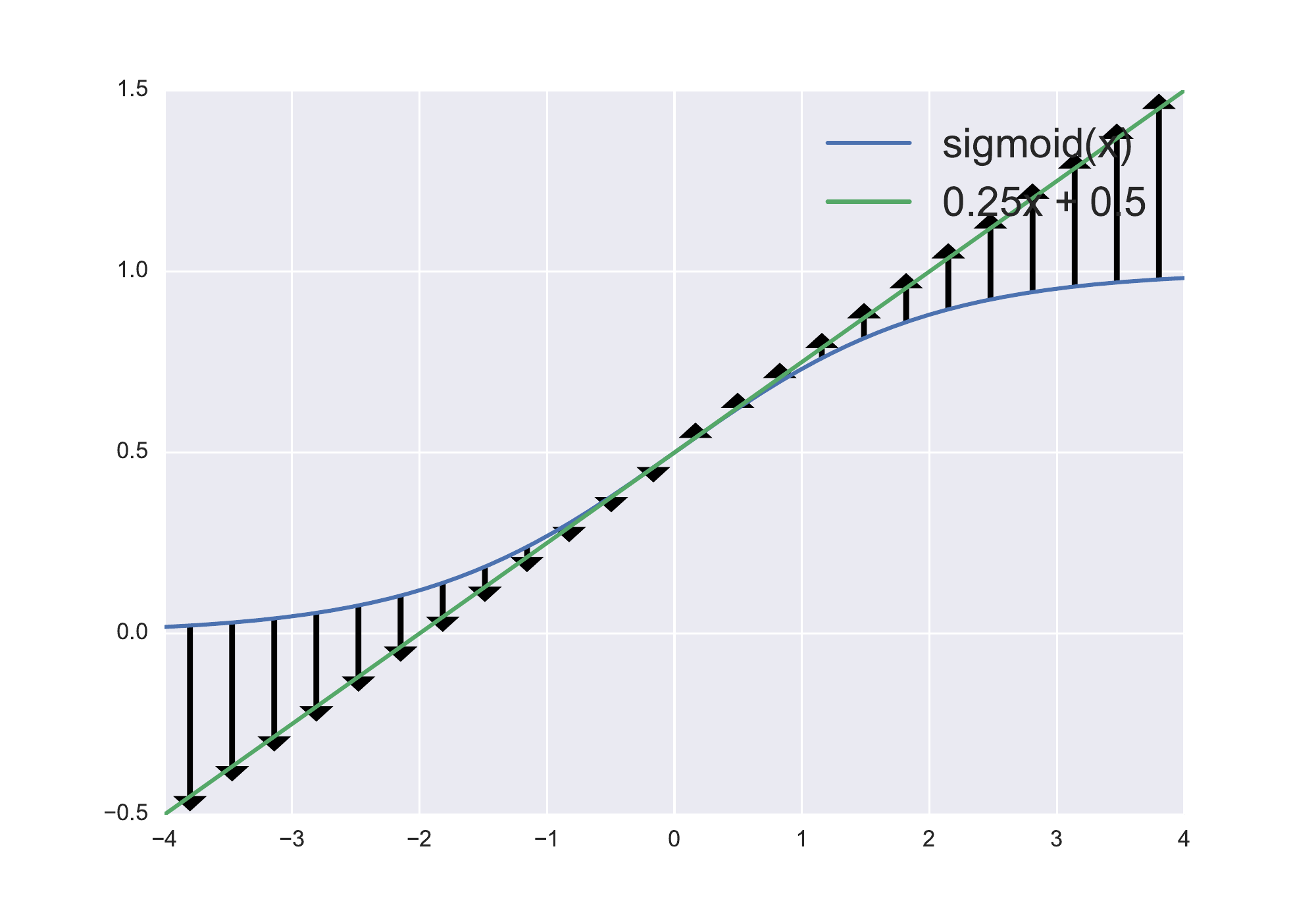}
        \caption*{b)}
        \label{fig:test2}
    \end{minipage}
    \captionof{figure}{
        The figures show how to evolve the model to make it closer to a linear
        network. Arrows denote the direction of the noise pushing the
        activation function towards the linear function.
        \textbf{a)} The quasi-convex envelope established by a
        $|\text{sigmoid}(\cdot)|$ around $|0.25x|$.
        \textbf{b)} A depiction of how the noise pushes the $\text{sigmoid}$ to
        become a linear function.
    } \label{fig:sigm_linsigm}
    \label{fig:linearizing_with_noise}
\end{figure}
The noise is sampled from a Normal distribution with mean $0$ and whose
standard deviation depends on c:
\[
    s_i \sim \mathcal{N}(0,~p~c~\sigma(x_i))
\]
The pseudo-code for the mollified activations is reported in Algorithm \ref{algo:moll_fn}.
\begin{algorithm}[h]
    \caption{Activation of a unit $i$ at layer $l$.}
    \label{algo:moll_fn}
    \begin{algorithmic}[1]
	    \State {$x_i \gets \vw_i^{\top} \vh^{l-1} + b_i$} 													\Comment{an affine transformation of $\vh^{l-1}$ }
	    \State $\Delta_i \gets \fu(x_i) - \f(x_i) $																\Comment{$\Delta_i$ is a measure of a saturation of a unit}
	    \State $\sigma(x_i) \gets (\text{sigmoid}(a_i \Delta_i) - 0.5)^2$                                   \Comment{std of the injected noise depends on $\Delta_i$}
	    \State $\xi_i \sim \mathcal{N}(0,~1)$																	\Comment{sampling the noise from a basic Normal distribution}
	    \State $s_i \gets ~p^{l}~c~\sigma(x_i)|\xi_i|$														\Comment{Half-Normal noise controlled by $\sigma(x_i)$, const. $c$ and prob-ty $p^l$}
        \State $\psi(x_i, \xi_i) \gets \sgn(\fu^{\ast}(x_i)) \text{min}(|\fu^{\ast}(x_i)|, ~|\f^{\ast}(x_i) + \sgn(\fu^{\ast}(x_i)) |s_i||) + \fu(0)$ 	\Comment{noisy activation}
	    \State $\pi^l_i \sim \text{Bin}(p^{l})$                               \Comment{$p^l$ controls the variance of the noise AND the prob of skipping a unit}
	    \State $\tilde{h}^{l}_i = \psi(x_i, \xi_i)$																\Comment{$\tilde{h}^{l}_i$ is a noisy activation candidate}
	    \State $\phi(\vh^{l-1}, \xi_i, \pi^l_i; \vw_i) = \pi^l_i h^{l-1}_i + (1 - \pi^l_i) \tilde{h}^{l}_i$ 	\Comment{make a HARD decision between $h^{l-1}_i$ and $\tilde{h}^{l}_i$}
    \end{algorithmic}
\end{algorithm}

\subsection{Mollifying LSTMs and GRUs}
In a similar vein it is possible to smooth the objective functions of LSTM and GRU networks by
starting the optimization procedure with a simpler objective function such as optimizing a
word2vec, BoW-LM or CRF objective function at the beginning of training and gradually
increasing the difficulty of the optimization by increasing the capacity of the network.

For GRUs we set the update gate to $\frac{1}{t}$ -- where $t$ is the annealing time-step -- and reset the gate to $1$ if
the noise is very large, using Algorithm \ref{algo:moll_fn}.
Similarly for LSTMs, we can set the output gate to $1$ and input gate to $\frac{1}{t}$ and forget
gate to $1-\frac{1}{t}$ if the noise is very large. The output gate is $1$
or close to $1$ when the noise is very large. This way the LSTM will behave
like a BOW model.
In order to achieve this behavior, the activations $\psi(x_t, \xi_i)$ of the gates can be formulated as:
$$\psi(x^l_t, \xi) = \f(x^l_t + p^l \sigmoid(x) |\xi|)$$
By using a particular formulation of $\sigmoid(x)$ that constraints it to be in
expectation over $\xi$ when $p^l=1$, we can obtain a function for $\gamma \in
\R$ within the range of $\f(\cdot)$ that is discrete in expectation, but still
per sample differentiable:

\begin{equation}
    \sigmoid(x^l_t) = \frac{\f^{-1}(\gamma) - x^l_t}{\E_{\xi}[|\xi|]}
    \label{eqn:gate_fn}
\end{equation}
We provide the derivation of Eqn. \ref{eqn:gate_fn} in Appendix \ref{sec:DerivGating}. The
gradient of the Eqn \ref{eqn:gate_fn} will be a Monte-Carlo approximation to the gradient of
$\f(\x^l_t)$.

\subsection{Annealing Schedule for $p$}

We used a different schedule for each layer of the network, such that the noise
in the lower layers will anneal faster. This is similar to the linearly decaying
probability of layers in \cite{DBLP:journals/corr/HuangSLSW16}. In our experiments,
we use an annealing schedule similar to inverse sigmoid rule in \cite{bengio2015scheduled}
with $p_t^l$,

\begin{equation}
    p^l_t = 1 -  \e^{-\frac{k \vv_t l}{tL}}
\end{equation}

with hyper-parameter $k \ge 0$ at $t^{th}$ update for the $l^{th}$ layer, where $L$ is the
number of layers of the model. We stop annealing when the expected depth
$p_t=\sum_{i=1}^{L}p_t^{l}$ reaches some threshold $\delta$. $\vv_t$ is a moving average of the
loss~\footnote{Depending on whether the model overfits or not, this can be a moving average of
training or validation loss.} of the network, therefore the behavior of the loss/optimization
can directly influence the annealing behavior of the network. Thus we will have:
\begin{equation}
    \lim_{\vv_t \rightarrow \infty} p_t^l = 1~~\text{and},~~\lim_{\vv_t \rightarrow 0} p_t^l = 0.
\end{equation}
This has the following desirable property: when the training-loss is high, the
noise injected into the system is large and the model is encouraged to do more
exploration, while when the model has converged the noise injected into the
system will be zero.

\section{Experiments}
In this section, we mainly focus on training of difficult to optimize models, in particular
deep MLPs with $\text{sigmoid}$ or $\text{tanh}$ activation functions. The details of the experimental
procedure is provided in Appendix \ref{sec:exp_details}.

\subsection{Deep MLP Experiments}
\paragraph{Deep Parity Experiments}

\begin{figure}[htp]
    \centering
    \begin{minipage}{.45\linewidth}
    \centering
    \includegraphics[width=\textwidth]{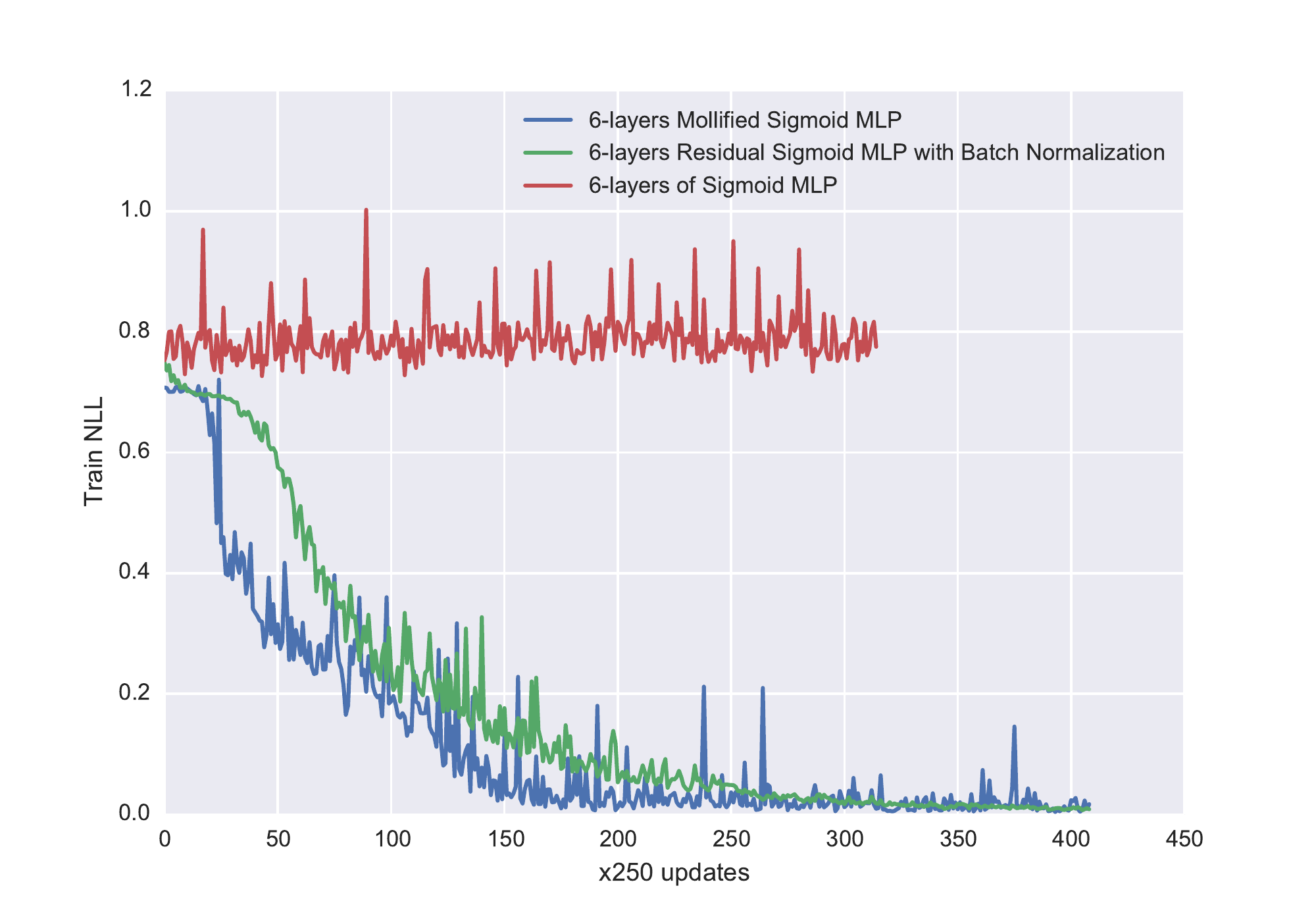}
    \captionof{figure}{The learning curves of a $6$-layers MLP with
        $\text{sigmoid}$ activation function on $40$ bit parity task.}
    \label{fig:deep_parity}
    \end{minipage}
    \hspace{12mm}
    \begin{minipage}{.45\linewidth}
    \includegraphics[width=\textwidth]{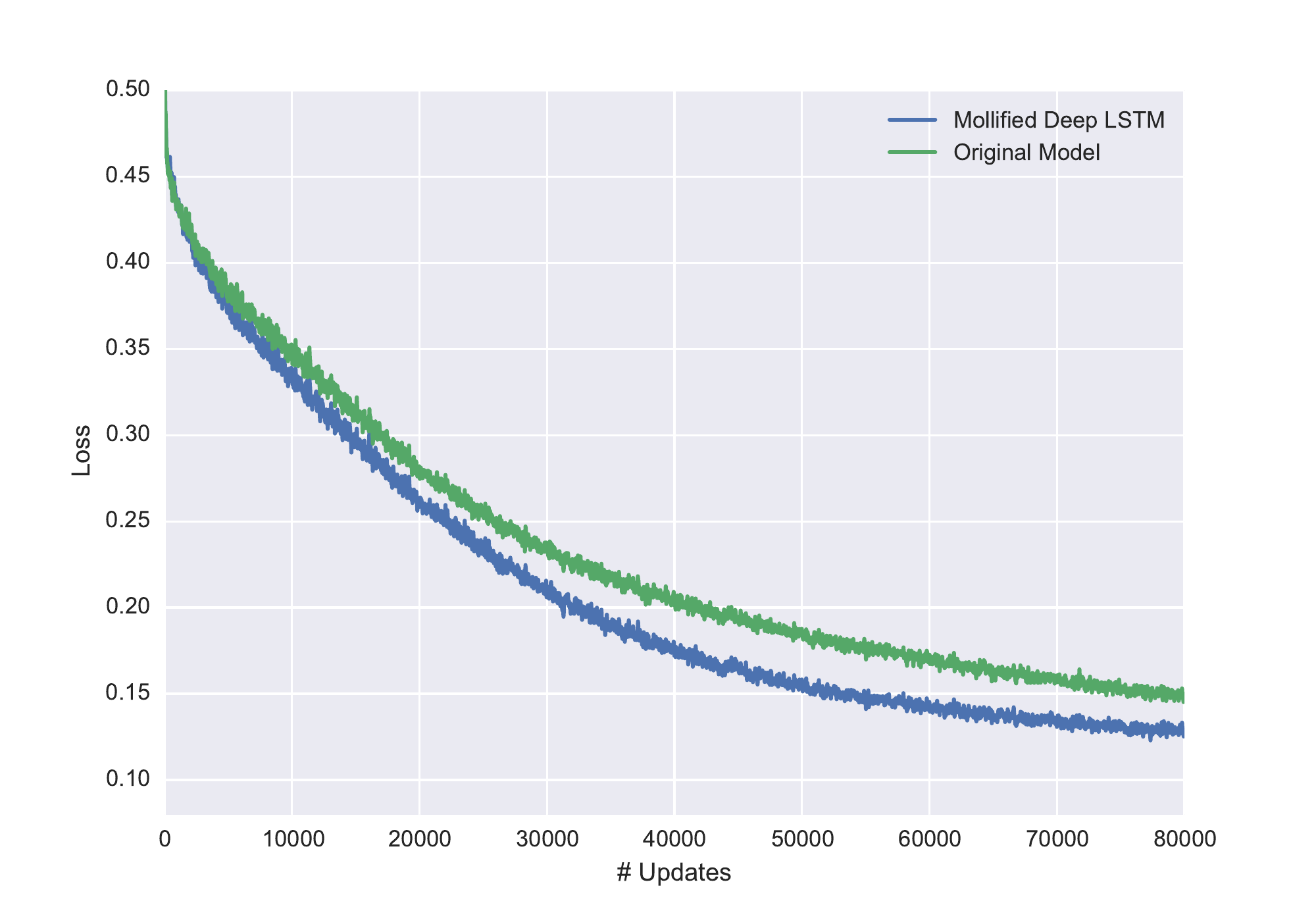}
    \caption{The training curve of a bidirectional-RNN that predicts the
        embedding corresponding to a sequence of characters.}
    \label{fig:deep_pento}
    \end{minipage}
\end{figure}

Training neural networks on a high-dimensional parity problem can be
challenging \citep{graves2016adaptive,kalchbrenner2015grid}. We experiment on
$40$-dimensional parity problem with $6$-layer MLP using sigmoid activation
function. All the models are initialized with Glorot initialization
\cite{glorot2011deep} and trained with SGD with momentum. We compare an MLP
with residual connections using batch normalization and a mollified network
with sigmoid activation function. As can be seen in Figure
\ref{fig:deep_parity}, the mollified network converges faster.
\vspace{-2mm}
\vspace{-2mm}



\paragraph{Deep Pentomino}
Pentomino is a toy-image dataset where each image has 3 Pentomino blocks. The
task is to predict whether if there is a different shape in the image or not
\citep{gulcehre2013knowledge}. The best reported result on this task with MLPs
is $68.15\%$ accuracy \citep{gulcehre2014learned}. The same model as ours
trained without noisy activation function and vanilla residual connections
scored $69.5\%$ accuracy, while our mollified version scored $75.15\%$ accuracy
after $100$ epochs of training on the $80k$ dataset.

\paragraph{CIFAR10}
We experimented with deep convolutional neural networks of 110-layers with
residual blocks and residual connections comparing our model against ResNet and
Stochastic depth. We adapted the hyperparameters of the Stochastic depth
network from \cite{huang2016deep} and we used the same hyperparameters for our
algorithm. We report the training and validation curves of the three models in
Figure~\ref{fig:sigm_linsigm} and the best test accuracy obtained early
stopping on validation accuracy over 500 epochs in Table
\ref{tbl:cifar10_deepconv}. Our model achieves better generalization than
ResNet. Stochastic depth achieves better generalization, but it might be
possible to combine both and obtain better results.

\begin{table}
    \centering
    \begin{minipage}[b]{.4\textwidth}\centering
        \begin{tabular}{@{}ll@{}}
            \toprule
            & Test Accuracy \\ \midrule
            Stochastic Depth & 93.25 \\
            Mollified Convnet & 92.45 \\
            ResNet & 91.78 \\ \bottomrule
        \end{tabular}
        \vspace{2mm}
        \caption{CIFAR10 deep convolutional neural network.}
        \label{tbl:cifar10_deepconv}
    \end{minipage}
    \hspace{5mm}
    \begin{minipage}[b]{.4\textwidth}\centering
        \begin{tabular}{@{}ll@{}}
            \toprule
             & Test PPL \\ \midrule
            LSTM & 128.4 \\
            Mollified LSTM & 123.6 \\ \bottomrule
        \end{tabular}
        \vspace{2mm}
        \caption{2-layered LSTM network on word-level language modeling for
            PTB.}
        \label{tbl:lstm_ptb}
    \end{minipage}
\end{table}
\vspace{-2mm}

\begin{figure}[htp]
    \centering
    \begin{minipage}{.4\textwidth}
       \centering
       \includegraphics[width=\textwidth]{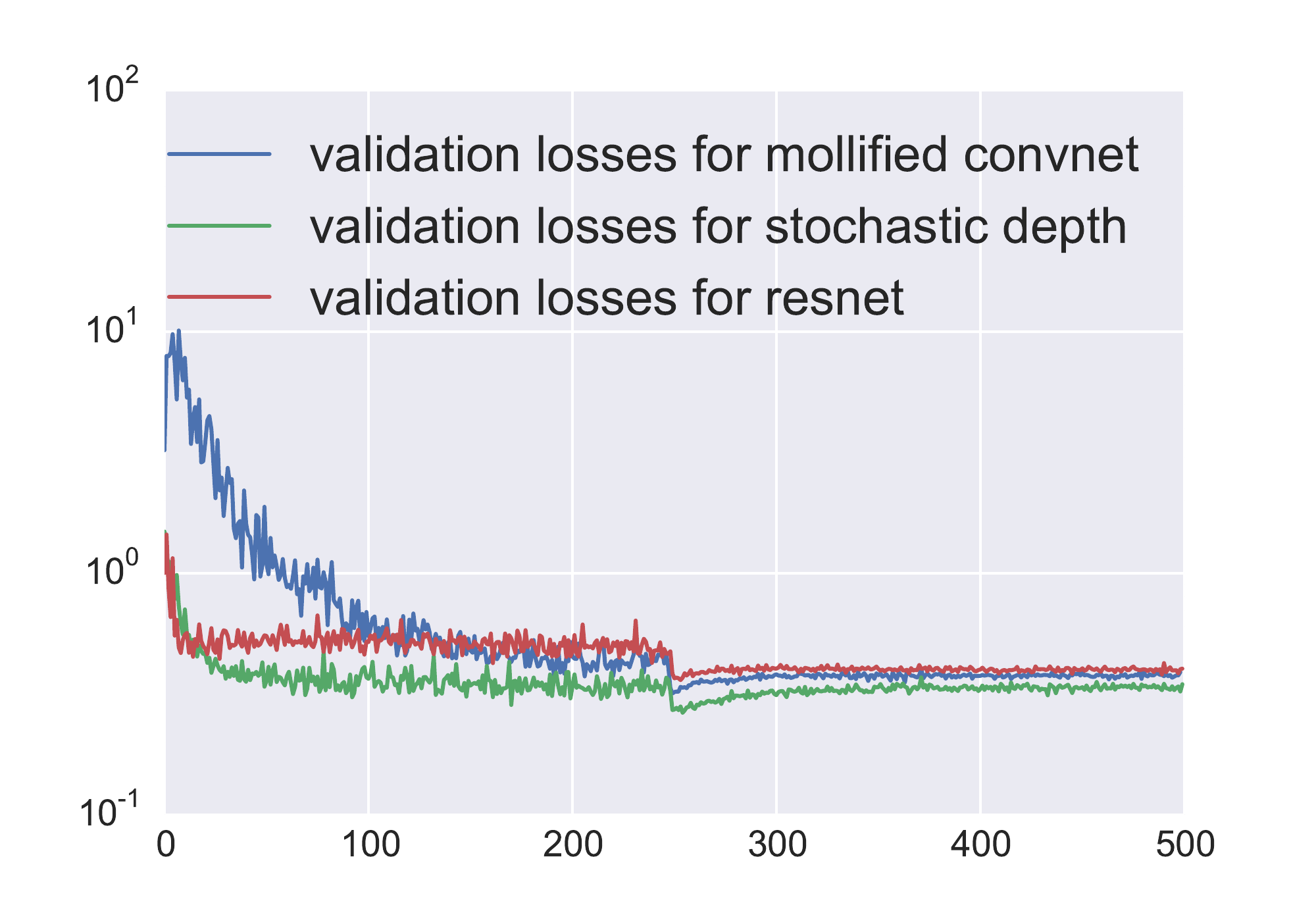}
       \vspace{-6mm}
       \caption*{a)}
       \label{fig:test1}
     \end{minipage}
     \hfil
     \begin{minipage}{.4\textwidth}
        \centering
        \includegraphics[width=\textwidth]{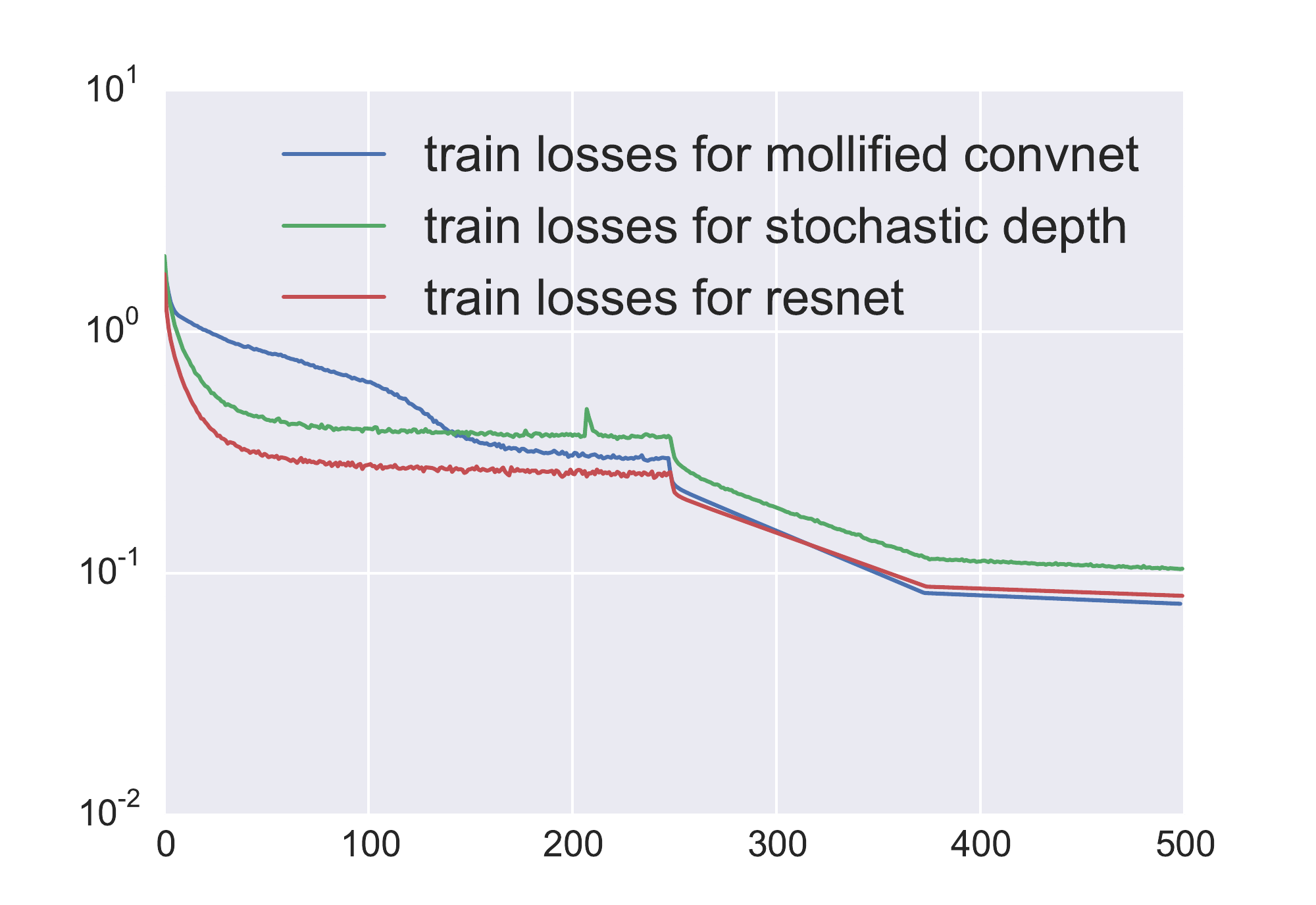}
        \vspace{-6mm}
        \caption*{b)}
        \label{fig:test2}
    \end{minipage}
    \captionof{figure}{
        Training and validation losses over 500 epochs of a mollified
        convolutional network composed by 110-layers. We compare against ResNet
        and Stochastic depth.
        } \label{fig:sigm_linsigm}
    \label{fig:linearizing_with_noise}
\end{figure}
\vspace{-2mm}

\subsection{LSTM Experiments}

\paragraph{Predicting the Character Embeddings from Characters}
Learning the mapping from sequences of characters to the word-embeddings is a difficult problem.
Thus one needs to use a highly non-linear function.
We trained a word2vec model on Wikipedia with embeddings of size $500$ \citep{mikolov2014word2vec} with a vocabulary of size
$374557$.

\paragraph{LSTM Language Modeling}
We evaluate our model on LSTM language modeling. Our baseline model is
a 2-layer stacked LSTM without any regularization. We observed that mollified model
converges faster and achieves better results. We provide the results for PTB
language modeling in Table \ref{tbl:lstm_ptb}.

\section{Conclusion}
We propose a novel method for training neural networks inspired by an idea of continuation, smoothing
techniques and recent advances in non-convex optimization algorithms.
The method makes the learning easier by starting from a simpler model
solving a well-behaved problem
and gradually transitioning to a more complicated setting.
We show improvements on very deep models, difficult to optimize tasks and compare with
powerful techniques such as batch-normalization and residual connections.

Our future work includes testing this method on large-scale language tasks that require
long training time, e.g., machine translation and language modeling. It is also intriguing to
understand how the generalization performance is affected by mollified networks, since the
noise injected during the training can act as a regularizer.

\bibliographystyle{natbib}
{\footnotesize
\bibliography{ml,refs,main,strings}}
\vspace{6mm}
\appendix
{\Large{\textbf{Appendix}}}
\section{Monte-Carlo Estimate of Mollification}
\label{sec:MonteCarloMolly}
\begin{equation}
         \label{eqn:mollifier_conv}
         \begin{aligned}
             \LL_K(\TT)~&=~(\LL \ast K)(\TT) ~=~\int_C \LL(\TT-\xi)K(\xi) d{\xi}~\text{which can be estimated by a Monte Carlo:} \\
                  & \approx \frac{1}{N} \sum_{i=1}^{N}
                  \LL(\TT-\xi^{(i)}),~\text{where}~\xi^{(i)}~\text{is a realization of the noise random variable } \xi\\
                  & \text{ yielding } \frac{\partial \LL_K(\TT)}{\partial \TT} \\
                  &\approx \frac{1}{N} \sum_{i=1}^{N} \frac{\partial \LL(\TT-\xi^{(i)})}{\partial \TT}.
         \end{aligned}
\end{equation}
Therefore introducing additive noise to the input of $\LL(\TT)$ is equivalent
to mollification.

\section{Linearizing ReLU Activation Function}
We have a simpler form of the equations to linearize ReLU activation function when $p^l
\rightarrow \infty$.
Instead of the complicated Eqn. \ref{eqn:psi_linear_conn_below}. We can use a
simpler equation as in Eqn \ref{eqn:simpler_molly_relu} to achieve the linearization
of the activation function when we have a very large noise in the activation function:

\begin{align}
    \label{eqn:simpler_molly_relu}
    s_i &= \text{minimum}(|x_i|, p \sigma(x_i)|\xi|)\\
    \psi(x_i, \xi_i, \vw_i) &= \f(x_i) - s_i
\end{align}


\section{Experimental Details}
\label{sec:exp_details}

\subsection{MNIST}
 The weights of the models are initialized with Glorot~\&~Bengio initialization~
 \cite{glorot2011deep}. We use the learning~rate of $4e-4$ along with RMSProp.
 We initialize $a_i$ parameters of mollified activation function by sampling it
 from a uniform distribution, $\text{U}[-2,2]$. We used $100$ hidden units at
 each layer with a minibatches of size $500$.

\subsection{Pentomino}
We train a $6-$layer MLP with sigmoid activation function using SGD and momentum. We used $200$ units
per layer with sigmoid activation functions. We use a learning rate of $1e-3$.

\subsection{CIFAR10}
We use the same model with the same hyperparameters for both ResNet, mollified network and the
stochastic depth. We borrowed the hyperparameters of the model from \cite{huang2016deep}. Our
mollified convnet model has residual connections coming from its layer below.

\subsection{Parity}
We use SGD with Nesterov momentum and initialize the weight matrices by using Glorot\&Bengio initialization\citet{glorot2011deep}.
For all models we use the learning rate of $1e-3$ and momentum of $0.92$.
$a_i$ parameters of mollified activation function are initialized by sampling from uniform distribution, $U[-2,2]$.

\subsection{LSTM Language Modeling}
We trained 2-layered LSTM language models on PTB word-level. We used the models with the same
hyperparameters as in \cite{zaremba2014learning}. We used the same hyperparameters for both the
mollified LSTM language model and the LSTM. We use hard-sigmoid activation function for both the
LSTM and mollified LSTM language model.

\subsection{Predicting the Character Embeddings from Characters}
We use $10k$ of these words as a validation and another $10k$ word embeddings as test set. We train a
bidirectional-LSTM on top of each sequence of characters for each word and on top of the representation
of bidirectional LSTM, we use a $5$-layered $\tanh$-MLP to predict the word-embedding.

We train our models using RMSProp and momentum with learning rate of $6e-4$ and momentum $0.92$. The size
of the minibatches, we used is $64$. As seen in Figure \ref{fig:deep_pento}, mollified LSTM network converges faster.

\section{Derivation of the Noisy Activations for the Gating}
\label{sec:DerivGating}
Assume that $z^l_t = x^l_t + p_t^l \sigmoid(x) |\xi^l_t|$ and $\E_{\xi}[\psi(x^l_t, \xi)] = t$.
Thus for all $z^l_t$,

\begin{align}
   \label{eqn:gen_in_eqn}
   \E_{\xi}[\psi(x^l_t, \xi^l_t)] &= \E_{\xi}[\f(z^l_t)],\\
   t &= \E_{\xi}[\f(z^l_t)],~\text{assuming}~\f(\cdot)~\text{behaves similar to a linear function}\\
   \E_{\xi}[\f(z^l_t)] &\approx \f(\E_{\xi}[z^l_t]) ~\text{since we use hard-sigmoid for}~\f(\cdot)~\text{this will hold.}\\
   \f^{-1}(t) &\approx \E_{\xi}[z^l_t] \\
\end{align}

As in Eqn. \ref{eqn:gen_in_eqn}, we can write the expectation of this equation as:
\[
    \f^{-1}(t) \approx x^l_t + p_t^l \sigmoid(x) \E_{\xi}[\xi^l_t]
\]
Corollary, the value that $\sigmoid(x^l_t)$ should take in expectation for $p_t^l = 1$ would be:
\[
    \sigmoid(x^l_t) \approx \frac{\f^{-1}(t) - x^l_t}{\E_{\xi}[\xi^l_t] }
\]

In our experiments for $\f(\cdot)$ we used the hard-sigmoid activation function. We used the
following piecewise activation function in order to use it as $\f^{-1}(x) = 4(x-0.5)$.
During inference we use the expected value of random variables $\pi$ and $\xi$.

\end{document}